# Variance of Twitter embeddings & Temporal Trends of Covid-19 Cases.

Mayank Sethi [+1], Ambika Sadhu [+2], Khushbu Pahwa [+3], Sargun Nagpal [4], Tavpritesh Sethi [*4]

1. Indraprastha Institute of Information Technology, Delhi, India
2. National Institute of Technology, Kurukshetra, India.
3. University of California, Los Angeles, California, USA.
4. Indraprastha Institute of Information Technology, Delhi, India

+Contributed Equally
*tavpriteshsethi@iiitd.ac.in

### Abstract

The ongoing COVID-19 pandemic necessitates agile surveillance and preparedness in the face of re-surging waves. India was particularly affected by the sudden onslaught of deaths caused by the Delta variant and threat for further waves is still looming. Here we seek to understand whether there was a statistical signal in social media that preceded the peak of cases during the second wave in India. This is motivated from the fact that the pandemic has led to an unprecedented usage of social media. However, evidence is shrouded within noise and this paper proposes a method for harnessing signals from Twitter that were indicative of upcoming surge in COVID-19 cases. We used time-series cross-correlation analysis between case numbers and a novel feature set from latent representation of COVID-19 Tweets in India. Statistically significant Tweet dimensions (STwD) with lead times of 15 and 30 days showed an $R^2$ scores of 0.67 and 0.47, respectively, indicating the potential of social media mining early signals for pandemic preparedness. Finally, we carried out an explainability analysis of STwDs to understand the thematic content of the tweets and discovered keywords that may be useful to track surges in the future.

*Keywords*: COVID-19, Social Media, Language Models, Time Series Prediction, Pandemic Preparedness.

## 1 Introduction

COVID-19 continues to spread in waves across all countries. On 11 March, 2020, WHO declared the infectious disease a pandemic. The pandemic has burdened the healthcare systems in almost every region, with a shortage in ICU beds, healthcare workers, and other resources such as PPE, ventilators, etc.[1] 2021 witnessed large-scale vaccination drives, which served as a critical tool in stabilizing the spread of the virus. However, some cases of vaccine breakthrough infections have also been reported [2]. As COVID-19 continues to mutate, multiple waves of cases with varying severity each time have become a cause of concern, requiring stringent measures. The virus leaves little time for effective decision-making with a rapid transmission speed. Though retrospective analysis plays a central role in governing administrative decisions relating to plans of action, prospective insights can aid in controlling the extent of loss of life and avoid an overwhelm of medical resources.

Since the outbreak began, user involvement in social media applications such as Twitter, WhatsApp, etc., has also increased enormously. These platforms serve as a medium for people to disseminate their opinions and, more importantly, voice concerns on the spread of the disease. The second wave in India, from April-June 2021, witnessed a surge in tweets related to fear of deaths due to the new strain, the crumbling capacity of hospitals, and the possibility of another lockdown[3]. Given that citizens are leveraging social media platforms, a pattern exists in the latent content and the corresponding rise in COVID-19. As hotspots emerged in India, Brazil, and the likes in 2020, most tweets were themed around the number of cases and deaths in an affected region[4]. This shows potential in using tweets as evidence for an upcoming surge. The semantic meaning of tweets can be exploited as a signal for the early prediction of a covid wave.

Several works have carried out a predictive modeling approach using different models in recent times. Some studies deployed linear regression models to predict the number of deaths [5], the number of confirmed cases using travel history data [6] as well as the number of recoveries [7].

Our approach captures the underlying relationship between these tweets and a covid surge for India. As concerns over newer waves arise, Twitter serves as a platform for changing emotions, indicative of a possible rise in cases. In recent times, word embeddings have proven to be effective in various text-related tasks as a learned representation of textual data as vectors. So, we leverage this representation of tweets to predict the number of cases in advance. With different leading periods, we train a predictive model using tweets and confirmed cases from August 2020 to July 2021. This monitoring of tweets to capture upcoming surges can be pivotal in governing administrative decisions and controlling the extent of the spread and the impact after that.

## 2 Data Collection and Pre-Processing

### 2.1 Datasets

For our use, the dataset consisting of tweets related to Covid-19 was downloaded from the publicly available repository maintained by Panacea Lab, Department of Computer Science, Georgia State University[8]. Apart from tweets, it also contains information about the country, its date, and time of origin. For this study, tweets pertaining to India were extracted between 1st August'20 to 31st July'21. The number of tweets for each day varies between 100-200 and 1500 or more per day, depending on the surge of Covid-19. Further, day-wise data of Covid-19 cases in India, from August 2020 to July 2021, was accessed from the repository maintained by Johns Hopkins University Center for Systems Science and Engineering (JHU CSSE) [9].

## 2.2 Data-Cleaning and Pre-Processing

Before using the tweets, some cleaning is performed on the text to convert it to a format suitable for modeling. The data cleaning step in the proposed pipeline consists of removing hashtags, mentions, URLs, and emojis. The cleaned data is passed through a Tokenizer that "splits" the phrase into "tokens". This step is followed by lemmatization (an evolution of stemming that reduces the word to its *lemma* or core representation). Lastly, we remove stopwords (excluding negation words), and punctuation removal, since these add only little meaning.

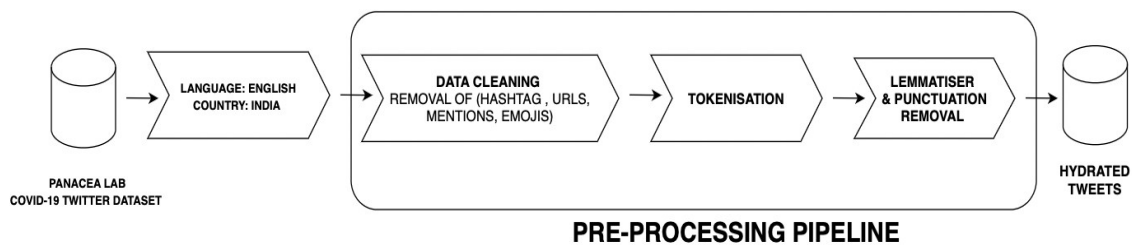

**Figure 1**: Pre-Processing of Tweets

## 2.3 Word Embeddings

Every "token" or "word" obtained after cleaning must be converted into a numerical representation. In this study, we leverage an unsupervised low-dimensional representation of words using the Word2Vec model [10]. The gensim [11] implementation of Word2Vec, with default parameters, was utilized. Each vector is learned from the tweet corpus with a window size of 5. Thus, we get a 100-Dimensional vector representation for every word in a tweet. To obtain the individual tweet vectors, we average out the vectors corresponding to each word present. Further, for attaining a single representation of each day, again, an average is taken over the vectors of all the tweets of that day. Thus, each day is represented as a 100-dimensional vector.

## 3 Feature Selection

Out of 100 dimensions of the day-wise vectors, *"Significant Dimensions"* or *SDs* are selected as the ones that show a "leading" relationship with the new case count. "Leading" means that these dimensions show a rise/fall before a certain period, post which the case count shows a change. Thus, temporal trend analysis of these SDs can facilitate early intervention for covid-spread control[1].

Two mechanisms were utilized for SD selection-

## 3.1 Cross-Correlation Analysis:-

Cross-correlation is a technique used to model the relationship between two-time series. This helps to capture the underlying trend between the time series of two variables. To determine the significant features having a leading trend with respect to the new cases, CCF (Cross-Correlation Function) has been used. Since the raw new-case series (response variable) is non-stationary. In order to perform the CCF analysis, it had to be first made stationary which entails that there should not be any change in its statistical properties over time. The stationarity of the new-case series was obtained on transformation employing second-order differentiation on the raw series. In order to make all the 100 dimensions of the tweet vectors stationary, rolling variance over 3 days has been used as the transformation. The raw and stationary new case series are as shown in Fig. 2.

---
[1] Dimension numbering is 0-indexed

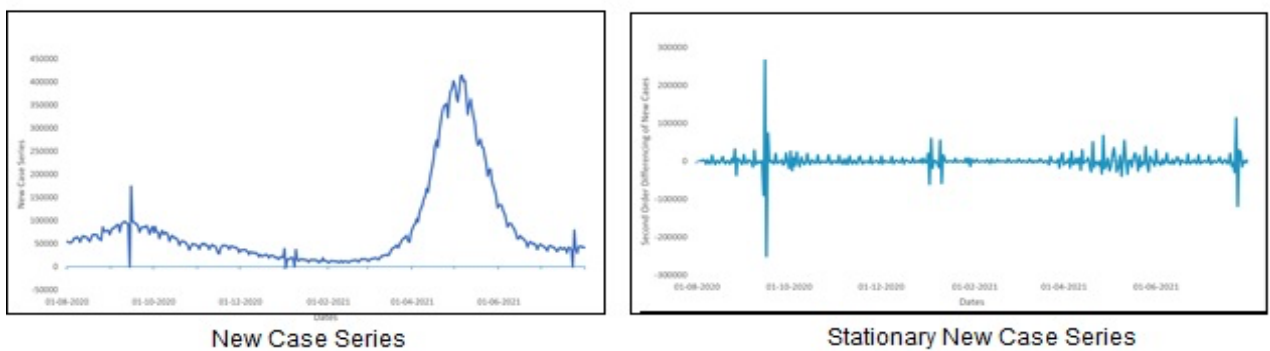

Figure 2. Covid Cases by Date

The CCF plots obtained for the cross-correlation between the dimensions and new-cases serves as a means to identify the lag of the dimensions (x-variable) that might be useful predictors for the new-cases (y-variable). 40 dimensions were identified as significant i.e. having a leading trend with respect to the new-cases. The CCF plot for one of the dimension is shown in Fig. 3.

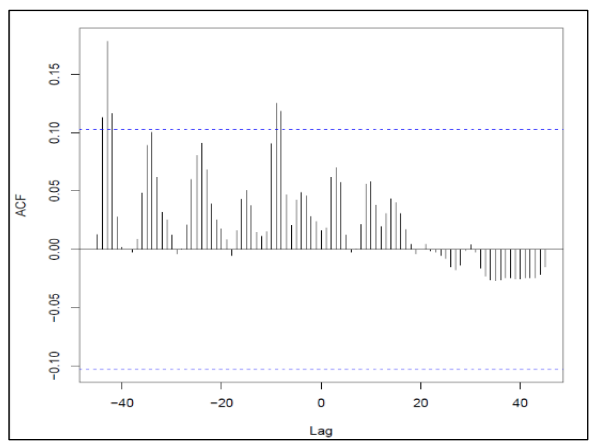

Figure 3: CCF Plot for Dimension 10

## 3.2 Boruta Analysis

Boruta Algorithm is a wrapper feature selection method that has been used for feature selection of the dimensions. Most of the traditional feature selection algorithms rely on a small subset of features which yields a minimal error on a chosen classifier. While fitting a random forest model on a data set, one can recursively get rid of features in each iteration which didn't perform well in the process. This will eventually lead to a minimal optimal subset of features as the method minimizes the error of random forest model. This happens by selecting an over-pruned version of the input data set, which in turn, throws away some relevant features. Boruta, on the other hand, finds all features which are either strongly or weakly relevant to the decision variable. Figure 4 shows the importance of dimensions as shown by Boruta.

---
**Algorithm 1: Boruta Algorithm**
---

1. Extend the information system by adding copies of all variables (the information system is always extended by at least 5 shadow attributes, even if the number of attributes in the original set is lower than 5).
2. Shuffle the added attributes to remove their correlations with the response.
3. Run a random forest classifier on the extended information system and gather the Z scores computed.
4. Find the maximum Z score among shadow attributes (MZSA), and then assign a hit to every attribute that scored better than MZSA.
5. For each attribute with undetermined importance perform a two-sided test of equality with the MZSA.
6. Deem the attributes which have importance significantly lower than MZSA as 'unimportant' and permanently remove them from the information system.
7. Remove all shadow attributes.
8. Repeat the procedure until the importance is assigned for all the attributes, or the algorithm has reached the previously set limit of the random forest runs.

---

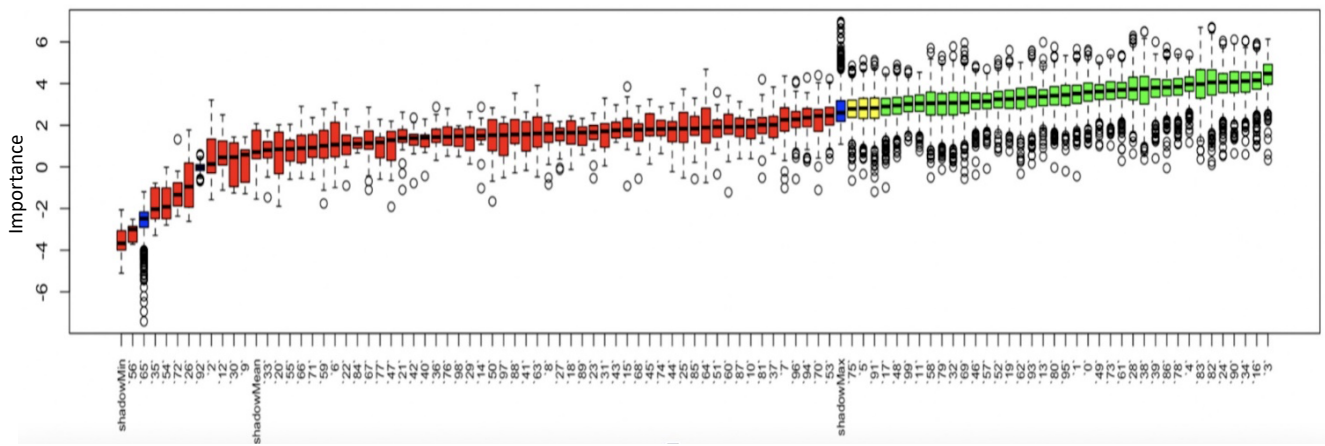

Figure 4: Importance of Dimensions from Boruta Analysis

## 4 Results

### 4.1 Model Development and Evaluation

Random Forest Regressor has been employed for modeling purposes. In our analysis, the response variable considered is the new case series, obtained by first-order differencing on the confirmed case series. The stationarity of the new case series is attained on second-order differencing with a P-value of $0.01 < 0.05$. The predictors are embedding dimension values, which attain stationarity on first-order differencing.

We considered varying parameters in terms of depth of the random forest as well as the number of estimators. Using these different values, two leading periods are tested- 30-day and 15 day. In case of 30 days, we use the dimension values to predict "new cases" after 30 days, i.e. a "shift" of 30 days. Similarly, 15-day lead time is used to predict new cases after a fortnight. Different models are trained using SDs from both CCF plots and Boruta, using the dimension values from 1 August to 15 April. R2 scores are calculated using the data from 1 August for the 15-day lag model.

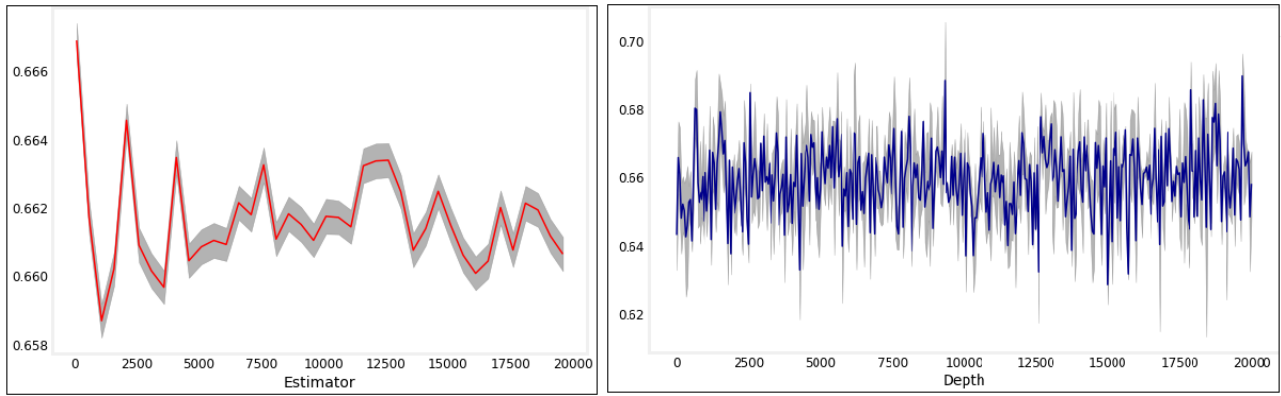

**Figure 5**: R-Square scores analysis for 15-day lead through CCF.

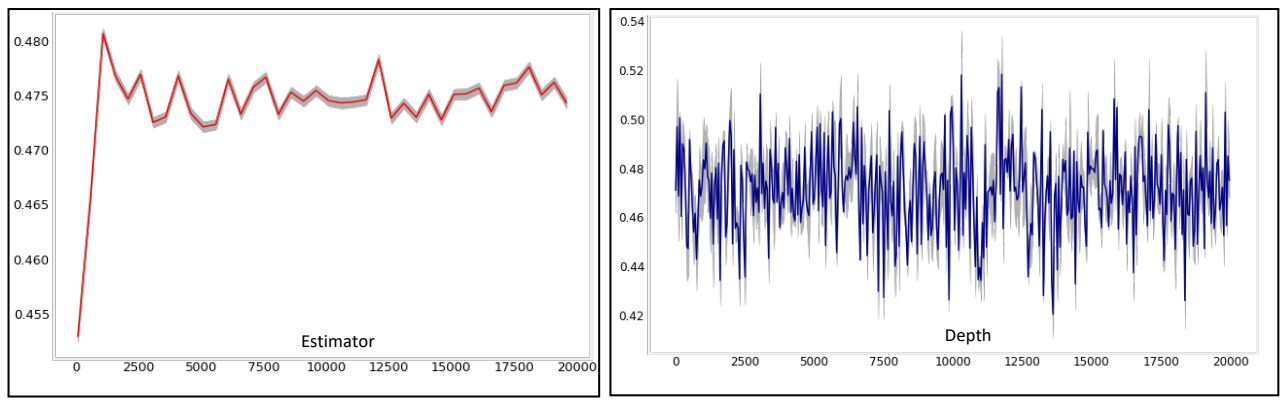

**Figure 6**: R-Square scores analysis for 15-day lead through Boruta Results.

## 4.2 Social Media trends and Tweet-based language modeling capture rapidly changing COVID-19 environment

As shown in the figure, the period of April-May 2021 saw a peak in tweets about covid, owing to the second wave. Likewise, as the cases started to reduce, the number of corresponding tweets also showed a similar decreasing trend. This confirms that the trends in usage of Twitter can capture the changing COVID scenario and their analysis can be useful in deriving insights. Further, visualization of word embeddings learned from tweets during this period show a similar pattern. t-SNE is a variation of the Stochastic Neighbour Embedding used primarily for visualizing high-dimensional data [12]. Since word embeddings are a geometrical representation of words in vector space, similar words are closer to each other. To visualize and interpret this similarity amongst the tweets, we used the TensorFlow Embedding Projector [13]. Figure 7 shows the t-SNE plot obtained for the day-wise averaged tweets. Here, each point in space represents one day. The cluster on left mostly contains dates from April to May, again corroborating the existence of a relationship between tweets and cases. Moreover, since word embeddings capture the semantic meaning of words in vector form, this clustering pattern indicates that there's a pattern associated with the type of tweets posted during this period. These visualizations serve as the basis for our hypotheses to use tweets for case count prediction.

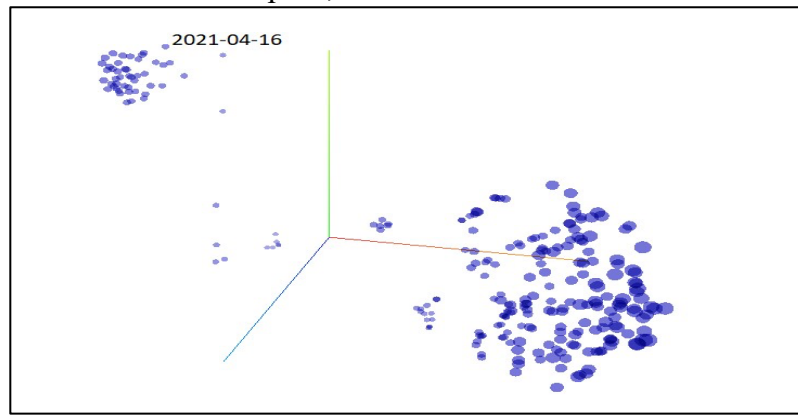

**Figure 7**: tSNE plot for Word2Vec vectors of Tweets

### 4.3 Temporal trends in daily tweets mimic Covid-19 surge

A deeper analysis of the specific tweets mentioned over the different time periods provides insights into the current public concerns of different terms in three separate phases- December-January-February (before the second wave), April-May (during the second wave) and June-July(post second wave). From Fig. 9, it can be seen that there is a clear variation in the occurrence of some key words pertaining to the pandemic and its management across the 3 periods. It can be validated that the words like "government", "cases", "lockdown", "reports", "mask", "help", "need", "pandemic" are higher in percentage for the April-May period when India witnessed the highest caseloads. National events such as the farmers' movement, election in several states etc. during the period led to an increase in the risk of transmission of COVID-19.

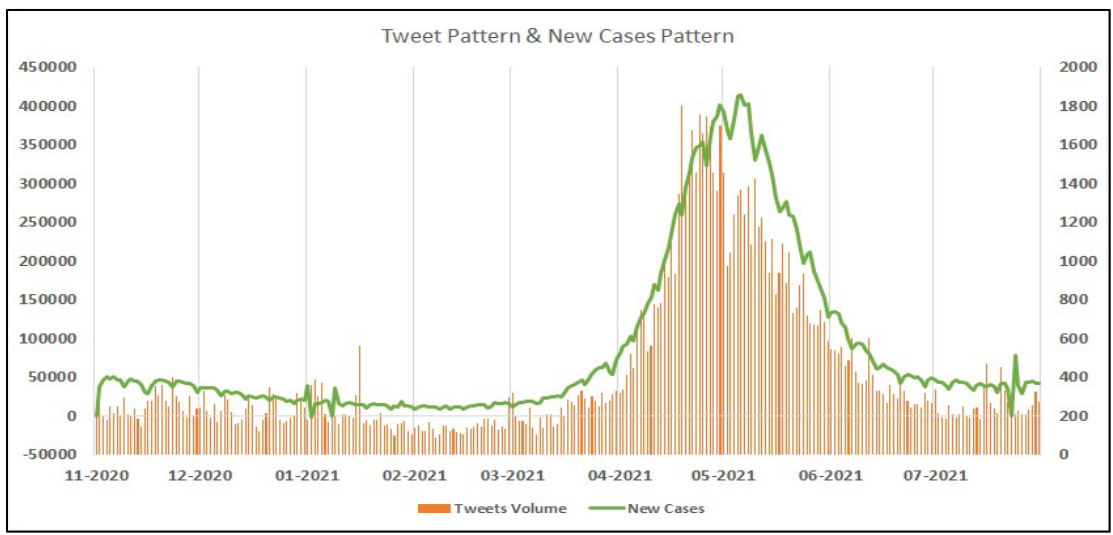

**Figure 8**: Tweets and New Case Trends

### 4.4 Rolling Variance in the latent space dimensions captures variability in the COVID-related tweets

In this study, first analysis was carried out on the line plots of the embedding dimensions with the daily new cases. Upon visual inspection, rolling variance over 3 days seemed to be a reasonable approximating feature. So, for the purpose of feature selection, the embedding dimensions were transformed using rolling variance. This was implemented using the pandas.core.window.rolling.Rolling.var package.

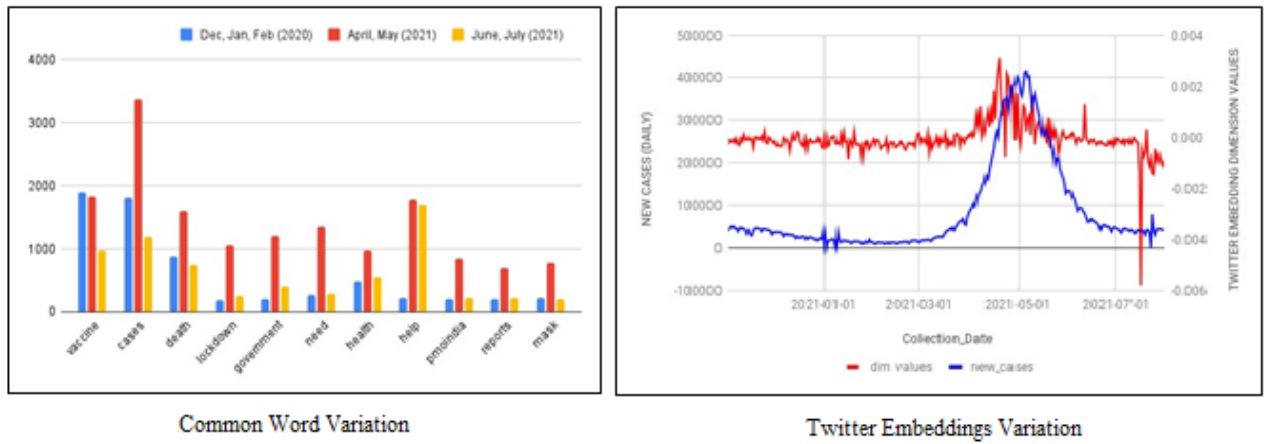

**Figure 9**: Variation of Words and Embeddings

### 4.5 Dimensions formed from Word Embedding are predictive of new COVID-19 caseloads

In Fig. 10., the 3-day moving average of the predicted case series using the dimensions from Cross-Correlation Analysis; and the actual case series are plotted. The blue lines represent the actual numbers during the decline of the second wave while the orange line represents the cases predicted by the model. The trend for predicted and actual caseloads are similar, and thus point out the effectiveness of the chosen twitter embedding dimensions for the prediction of caseloads.

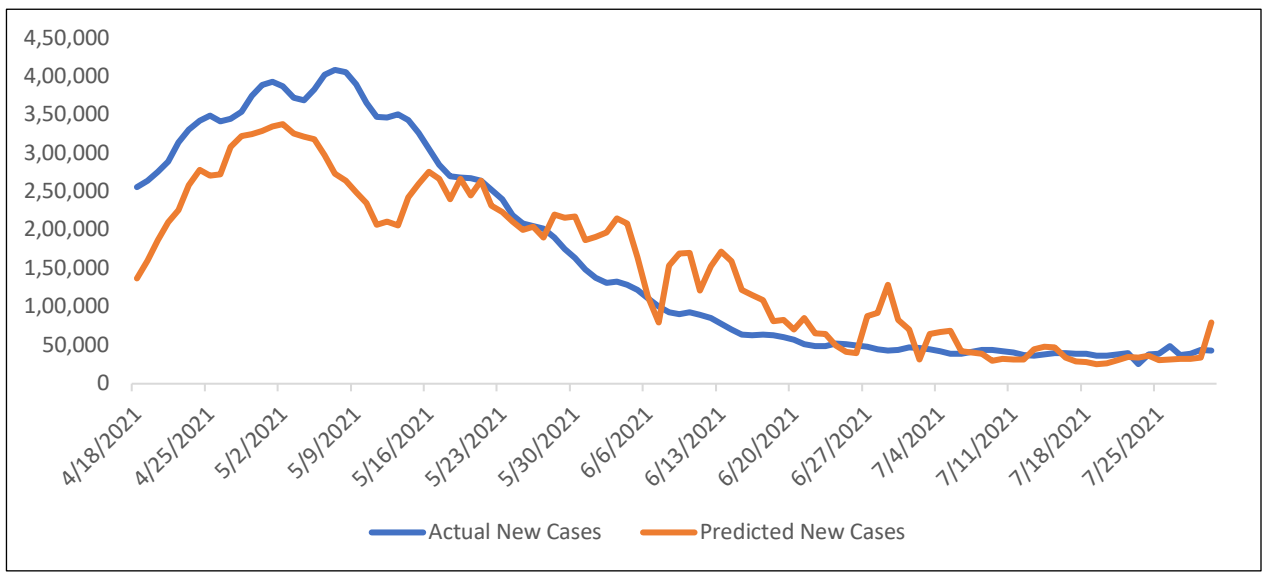

**Figure 10**: 3-day Moving-Average of Predicted and Actual Case loads

### 4.6 Each specific dimension from the SDs represent certain themes of the scenario underway.

In the table below, each dimension has few key words associated to it with highest weights for the specific dimension. Upon studying these keywords for few of the dimensions, we see that they tend to represent certain message of the situation of Covid-19 spread in the geography. Considering the effectiveness of this method to give hint of future scenarios, we can check the type of keywords occurring the tweets to catch certain issue, or a theme being raised by the people over the platform.

**Table 1**: Themes represented by SD:-

| Dimension | Selected Keywords | Key Inferences |
|---|---|---|
| 3 | "government", "leader", "political", "vaccination", "failure", "shocking" | Administrative inefficiency in handling COVID surge, at both national and local levels. |
| 24 | "hope", "calm", "together", "fight", "stay at home" | Optimism to fight the pandemic- General public, as well as political leaders, tweeted about the need to stay hopeful and calm in order to fight COVID-19. |
| 69 | "enforcement", "protocols", "high court", "lockdown" | Protocol enforcement to handle the situation. |
| 37 | "Decision", "Lockdown", "Election", "Adani", "Ambani", "Oxygen plant" | Public opinion on requirement of oxygen plants, lockdown, and need for public figures to come forward. |
| 14,26 | "ventilator", "ICU", "beds", "tocilizumab", "oxygen" | Request for resources as their availability declined with the rising cases. |

## 5 Discussion

Throughout the pandemic, researchers from across the world developed novel research methodologies for the spatiotemporal predictions of COVID-19 cases using both statistical and deep learning techniques. Some studies relied on clinical data [14] relied on data from clinical records of patients, to predict the risk of COVID-19 progression. However, gathering clinical data is a difficult task. On the other hand, some works focused on symptoms mentioned in tweets and google searches [16], thus establishing the utility of social media as a signaling mechanism.

By using word embeddings to represent tweets in this work, we rely on the ability of word vectors to capture varying semantic meaning, with changing context of the words in the individual tweets. Clustering patterns in the tSNE visualizations confirm the presence of temporal trends in the tweets.

In this pipeline, vectors for each day have been obtained by averaging the individual tweet vectors. As indicated by cross-correlation analysis, values of the vector dimensions depict a relationship with COVID cases. Given that vector values change rapidly, 3-day rolling variance appeared to be a good feature. To carry forth a "prospective" analysis, we select dimensions(SDs) that show a leading relationship with the new cases, both based on cross-correlation as well as Boruta. The corresponding R2 scores from the Random Forest Regressor highlight the utility of leveraging tweet embeddings as a surveillance mechanism and validate the selection of SDs. Although exact values may not match, merely a rising/falling trend in an upcoming time window can help govern policy making.

Corresponding to every dimension, there are tweets that contribute the highest to it. This is identified using the absolute value of tweet vector along that dimension. Thus, for every dimension, tweet vectors are sorted based on the absolute value of the vector, and a subset of top 30 tweets was selected. Table 1 lists the key messages highlighted by 6 SDs, shortlisted from dimensions common to both CCF and Boruta analysis. The ability of vector dimensions to capture specific aspects is demonstrated by the tokens discovered in SDs. The selected SDs are able to portray public emotions over the pandemic, government actions, demands for resources etc. These insights coupled with spatial data can be harnessed for tracking the situation in different areas. Some topics of discussion, that were trending around the second wave, were not discovered in the SDs. These include "Kumbh Mela", the religious festival that led to a huge number of devotees overcrowding the venue. This Mela became a point of controversy for becoming a leading cause for the second wave. However, none of the top tweets mentioned specifically the mela. This can be attributed to the averaging being done to obtain the vectors. Similarly, the SDs do not indicate any mention of *mucormycosis* or black fungus. But, since the incidence of these cases was also very low, the number of people who tweeted about it is also significantly less.

Thus, our contributions in this work are twofold. First, it paves way for research towards leveraging social media as a 15-day lead-tracker for Covid-19, which can enable timely interventions for better disease control. Though the exact number of new cases may not match, a rising/falling trend can provide a prospective analysis as to how the situation is going to change. Secondly, it introduces *Significant Dimensions (SDs)* and associated themes. Analyzing the messages associated with these SDs can aid in selecting the domains where administrative decisions could be channelled. These themes can be coupled with locational data, to analyze spatial trends, similar to [15].

However, there are several limitations to this study. First, we have leveraged Word2Vec with default parameters for generating word embeddings. Given that the vector for a day is an average of the tweets for that day, there is an associated loss of context. A comparison with other language models such as BERT would enable selecting the one that gives the best results. Further, instead of averaging the individual tweets, learning day-wise embeddings could capture context better, and provide a deeper understanding of thematic distribution across dimensions. Altering the parameters for Word2Vec is also a possible extension of this study.

Currently, the study is limited to India. Covering more countries across the world can help in identifying "global" dimensions, whose tracking can lead to a more generic analysis. Moreover, the results are limited to English language only. Including vernacular languages can support a "local" analysis as well. The results obtained are also limited by the number of active twitter users, as well as users who actually tweet about the pandemic. Lastly, current study uses the rolling variance of raw values of the dimensions for prediction. A future extension could be to explore a different feature, such as entropy, or combining a set of features.

**Acknowledgments**

This work was supported by the Delhi Cluster- Delhi Research Implementation and Innovation (DRIIV) Project supported by the Principal Scientific Advisor Office, Prn. SA/Delhi/Hub/2018(C) and the Center of Excellence in Healthcare supported by Delhi Knowledge Development Foundation(DKDF) at IIIT-Delhi. We also thank Ridam Pal and Aditya Nagori for their valuable inputs.